\documentclass[sigconf]{acmart}
\AtBeginDocument{%
  \providecommand\BibTeX{{%
    \normalfont B\kern-0.5em{\scshape i\kern-0.25em b}\kern-0.8em\TeX}}}

\setcopyright{acmcopyright}
\copyrightyear{2022}
\acmYear{2022}
\acmDOI{XXXXXXX.XXXXXXX}

\acmConference[ICAIF 2022 Workshop]{}{Nov 02, 2022}{New York, NY}

\begin{document}

\title{Rule-adhering synthetic data - the lingua franca of learning}

\author{Michael Platzer}
\email{michael.platzer@mostly.ai}
\affiliation{%
  \institution{MOSTLY AI}
  \city{Vienna}
  \country{Austria}
}

\author{Ivona Krchova}
\email{ivona.krchova@mostly.ai}
\affiliation{%
  \institution{MOSTLY AI}
  \city{Vienna}
  \country{Austria}
}

\begin{abstract}
AI-generated synthetic data allows to distill the general patterns of existing data, that can then be shared safely as granular-level representative, yet novel data samples within the original semantics. In this work we explore approaches of incorporating domain expertise into the data synthesis, to have the statistical properties as well as pre-existing domain knowledge of rules be represented. The resulting synthetic data generator, that can be probed for any number of new samples, can then serve as a common source of intelligence, as a lingua franca of learning, consumable by humans and machines alike. We demonstrate the concept for a publicly available data set, and evaluate its benefits via descriptive analysis as well as a downstream ML model.
\end{abstract}




\maketitle

\section{Motivation}

AI has progressed over the past decades from symbolic AI towards a far more capable subsymbolic AI. Whereas the former feeds upon existing domain knowledge on rules and associations, the latter is taught to learn these "on its own" from existing data. The increase in data, in compute and in model capacity have led to significant breakthroughs, and already allow to exceed human performance on a wide variety of tasks. Analogously, synthetic data has drastically advanced over the past couple of years. While rule-based fake data generators can create an arbitrary number of records, these will only represent the limited knowledge of an expert, that encodes the underlying rules. On the other hand, generative deep neural networks allow to simulate statistically representative, highly realistic, yet truly novel records in a fully automated fashion\cite{karras2017progressive, brown2020language, platzer2021holdout}. This new breed of AI-powered synthetic data generators is built to distill the insights of existing data, to detect rules and associations that generalize beyond the individual record, to then encode the retained information in the form of new data samples. And with the synthetic data being represented with the same semantics as the original data, this intelligence can be consumed, processed and learned from by people of all backgrounds (policy makers, business managers, data scientists, end users, advocacy groups, etc.). E.g., a fictional record that represents the case of a 38-year old single mother with 3 children, who earns a minimum salary while she pays back an outstanding housing loan of \$120k, can easily be reasoned upon with and without an education in statistics. Thus, synthetic data can serve as a \textit{lingua franca}, as a common language, of learning, for humans as well as for machines alike, as it allows to share the patterns and the diversity of a population via an unlimited number of representative samples, all without infringing anyone's privacy.

A key emerging trend in AI is the marriage of symbolic and subsymbolic AI into a hybrid neurosymbolic AI, that makes best use of both approaches, by learning from existing data as well as from existing domain expertise. It is motivated by the need to become more data efficient, and to generalize well into sparse areas of the data space. In particular, for rare yet high impact cases, like fraud, defects or accidents, practitioners seek to gain more confidence in spotting and understanding these. In a similar vein, we demonstrate with this paper the concept of \textit{rule-adhering synthetic data}. AI-generated synthetic data that can both learn from data as well as from rules, and can thus provide insights and confidence with respect to otherwise sparsely populated data regions.

\section{Concept}

Domain expertise can take various forms, ranging from exact relations ($a + b = c$), to constraints ($\texttt{deathYear} >= \texttt{birthYear}$), to priors ($x \sim N(0,1)$), and so forth. In this work we focus on hard constraints for categorical attributes, where a domain expert knows that certain combinations of attribute values are impossible to occur. Given sufficient amount of training data, a machine learning model is expected to pick up the non-occurrence of such combinations, and thus assign already a low probability to it. However, without any additional information, neither a human nor the learning algorithm can conclude whether a given combination is indeed impossible, or just so rare, that it hasn't yet been observed in the training data.

In order to incorporate the knowledge on impossible combinations into synthetic data, we pursue two approaches:
\begin{enumerate}
    \item Rules for training: add a rule-specific training loss component to penalize assigning any non-zero probability to invalid areas of the data space\footnote{See \cite{seo2021controlling} or \cite{tiwald2021representative} for models, where additional loss components are being added to the training objective.}.
    \item Rules for sampling: set the probability of invalid areas of the data space to zero during the sampling phase\footnote{See \cite{holtzman2019curious} for an example, where probabilities of a generative model are being manipulated during sampling.}.
\end{enumerate}

While the former impacts the learning of the generative model, the latter only applies to the sampling from an already trained model. On the other hand, while the former reduces the likelihood of invalid records, which itself depends on the relative weight of the additional loss component, it is only the latter that strictly rules these out.

We explore both approaches, as well as a combination of these. We do so by extending the underlying model of a commercial solution provider for structured synthetic data\footnote{MOSTLY AI, \href{https://mostly.ai/}{https://mostly.ai/}}. All generated data sets, as well as the corresponding analysis, are made available at \href{https://github.com/mostly-ai/paper-rule-adherence}{https://github.com/mostly-ai/paper-rule-adherence}.

\section{Demonstration}

\begin{figure}[t]
  \centering
  \includegraphics[width=\linewidth]{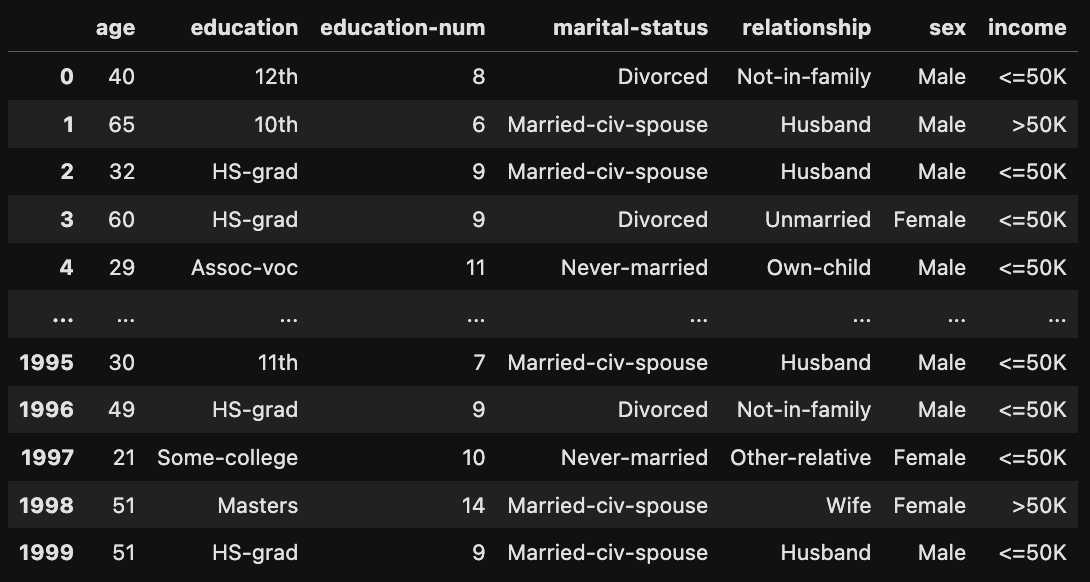}
  \caption{Sample records for the Census Income data set}
  \label{fig:adult_sample}
\end{figure}

To demonstrate the aforementioned approaches we will generate rule-adhering synthetic versions of the \textit{Census Income}, also known as \textit{adult}, data set, that is obtained from the UCI Machine Learning repository \cite{uci}. This is a widely studied, tabular data set, consisting of 48,842 records across a range of mixed-type variables. Figure~\ref{fig:adult_sample} shows a selection of attributes for a random sample of records.

Exploring the data set reveals a handful of rules / constraints, that seem to be in place:
\begin{enumerate}
    \item If the \textit{education} level is either \textit{Doctorate} or \textit{Prof-school}, then the person is at least 25 years old.
    \item If the relationship status is either \textit{Husband} or \textit{Wife}, then the marital status is always \textit{Married-civ-spouse}. On the other hand, if the marital status is \textit{Married-civ-spouse}, then the relationship status cannot be \textit{Unmarried}.
    \item The attributes \textit{education} and \textit{education-num} exhibit a strict 1:1 relationship and thus represent identical information.
    \item If \textit{relationship} is \textit{Wife}, then the person is \textit{female}, if it is \textit{Husband} then the \textit{sex} is \textit{male}\footnote{Actually, this "rule" is violated in 4 cases within the original data set. Hence, male wives and female husbands are seemingly possible, yet very very rare. Whether they then occur or not, then highly depends on the size of the actual sample that is being made available.}.
\end{enumerate}

Let's consider that rules (1), (2), (3) and (4) are all valid, and shall be respected within the generated synthetic data. Let's further assume that we only have access to a small subset of 2,000 records. In order to share the knowledge of this data as well as the existing rules, we proceed with training a generative model on top of these samples, to then create 100,000 novel, rule-adhering synthetic records. For comparison, we do this once by incorporating rules into the learning phase, once into the sampling phase, and once by combining both of the approaches at the same time. In addition, we compare this to a synthetic data set, that is being generated without any explicit knowledge of these four rules.

\begin{figure}[t]
  \centering
  \includegraphics[width=\linewidth]{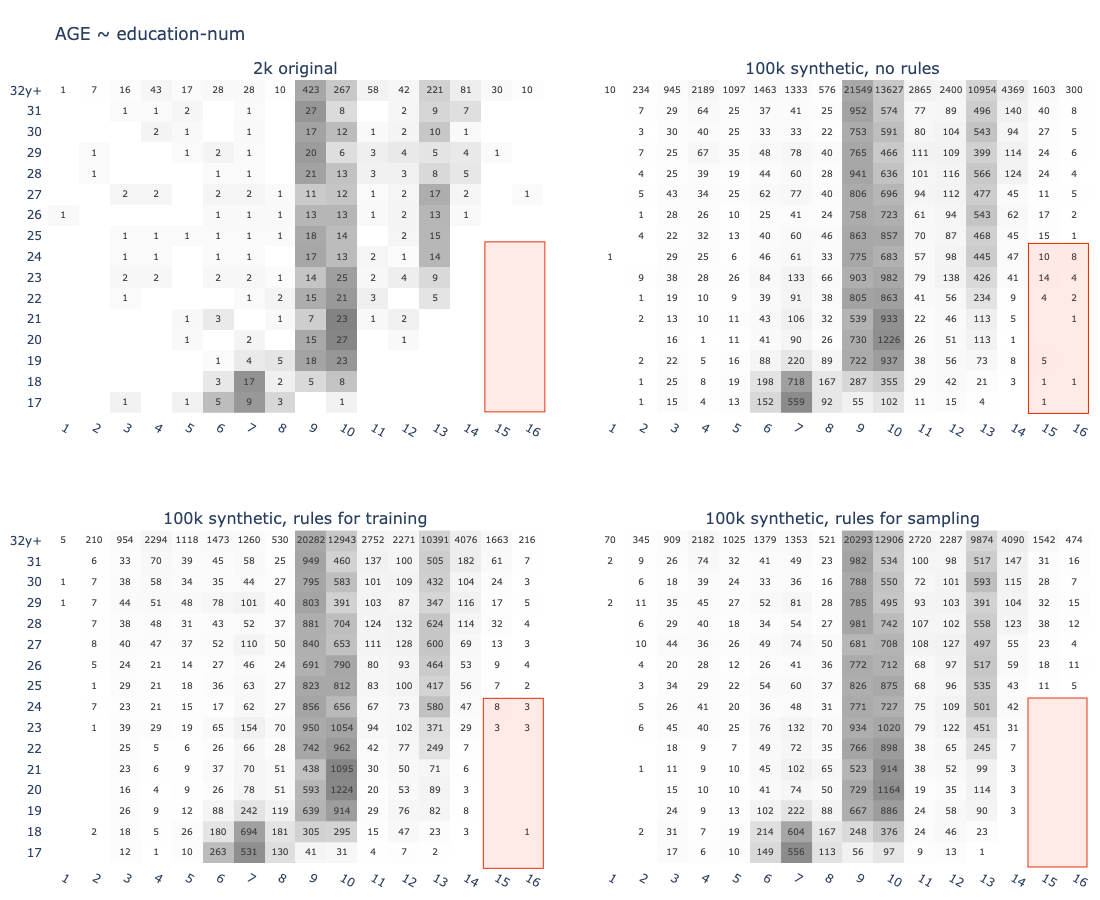}
  \caption{Rule (1) for \textit{age} vs. \textit{education}}
  \label{fig:adult_age}
\end{figure}

\begin{figure}[t]
  \centering
  \includegraphics[width=\linewidth]{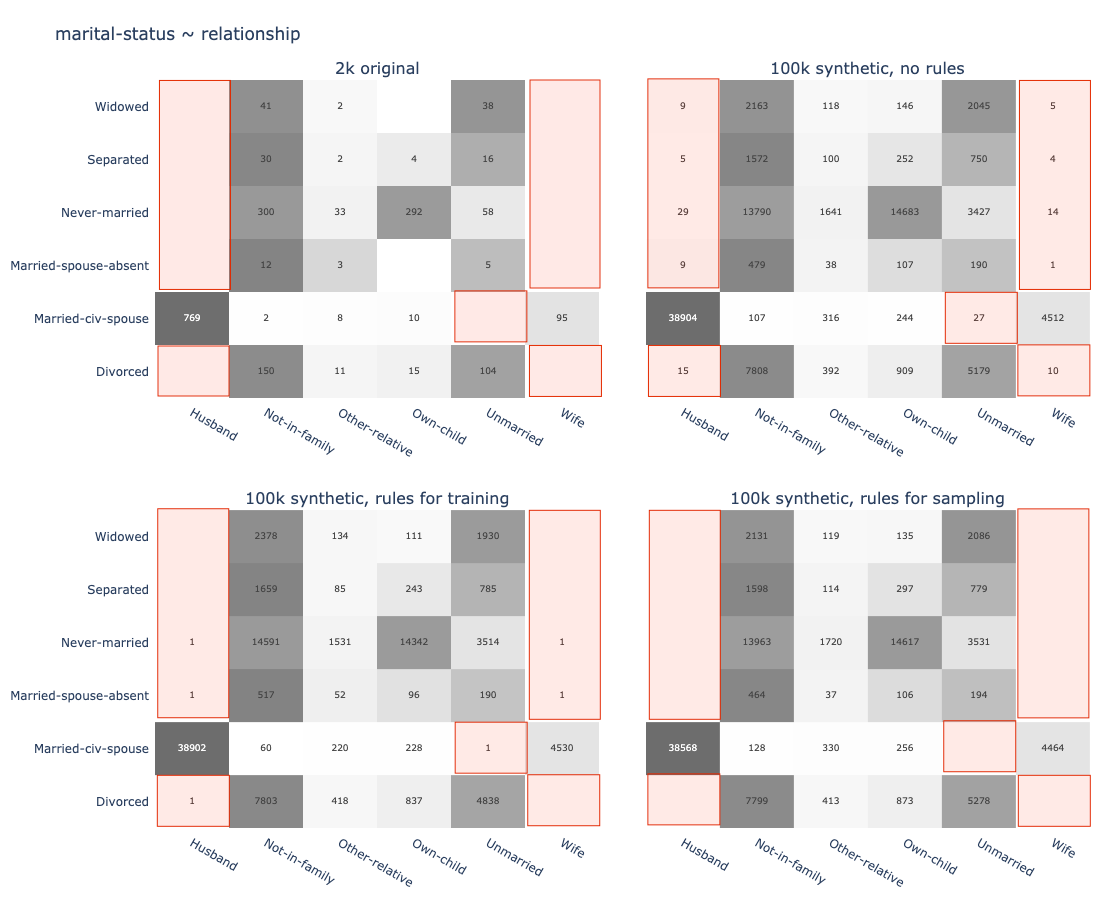}
  \caption{Rule (2) for \textit{marital-status} vs. \textit{relationship}}
  \label{fig:adult_relationship}
\end{figure}

\begin{figure}[t]
  \centering
  \includegraphics[width=\linewidth]{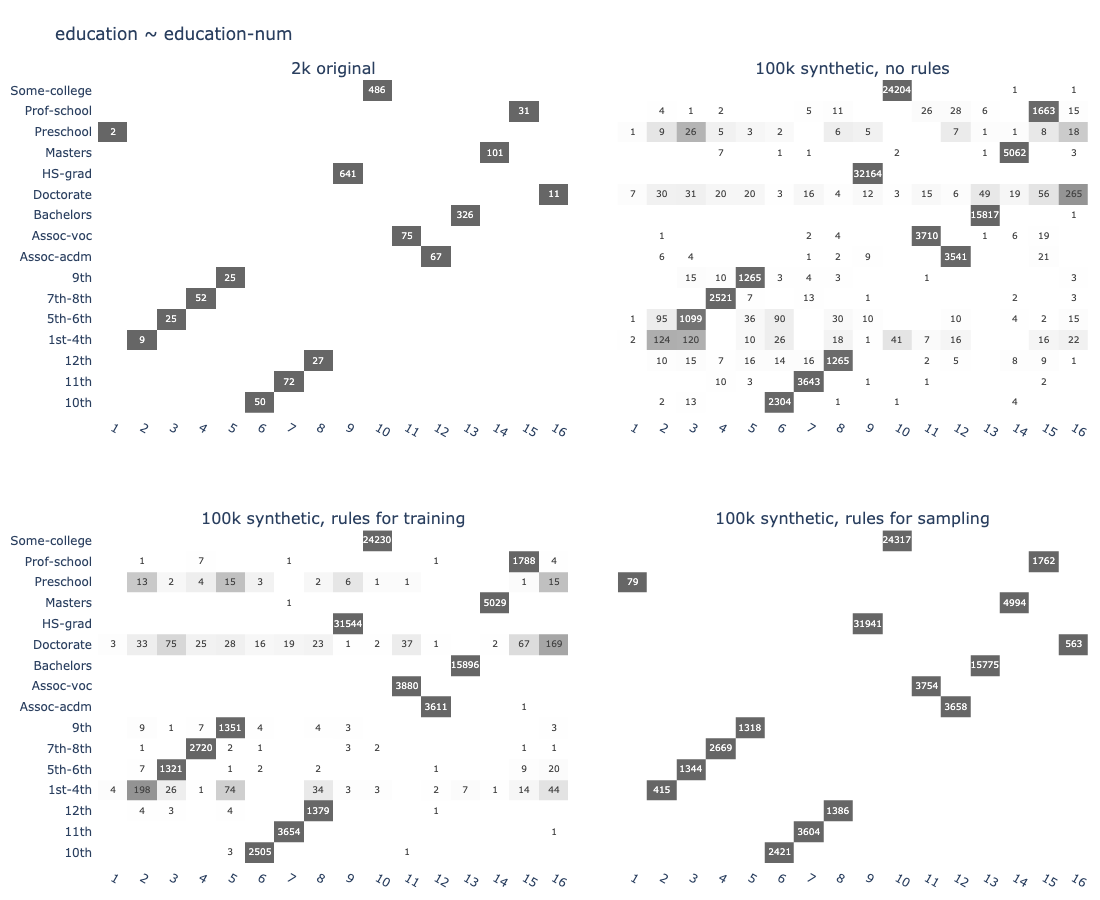}
  \caption{Rule (3) for \textit{education} vs. \textit{education-num}}
  \label{fig:adult_education}
\end{figure}

\begin{figure}[t]
  \centering
  \includegraphics[width=\linewidth]{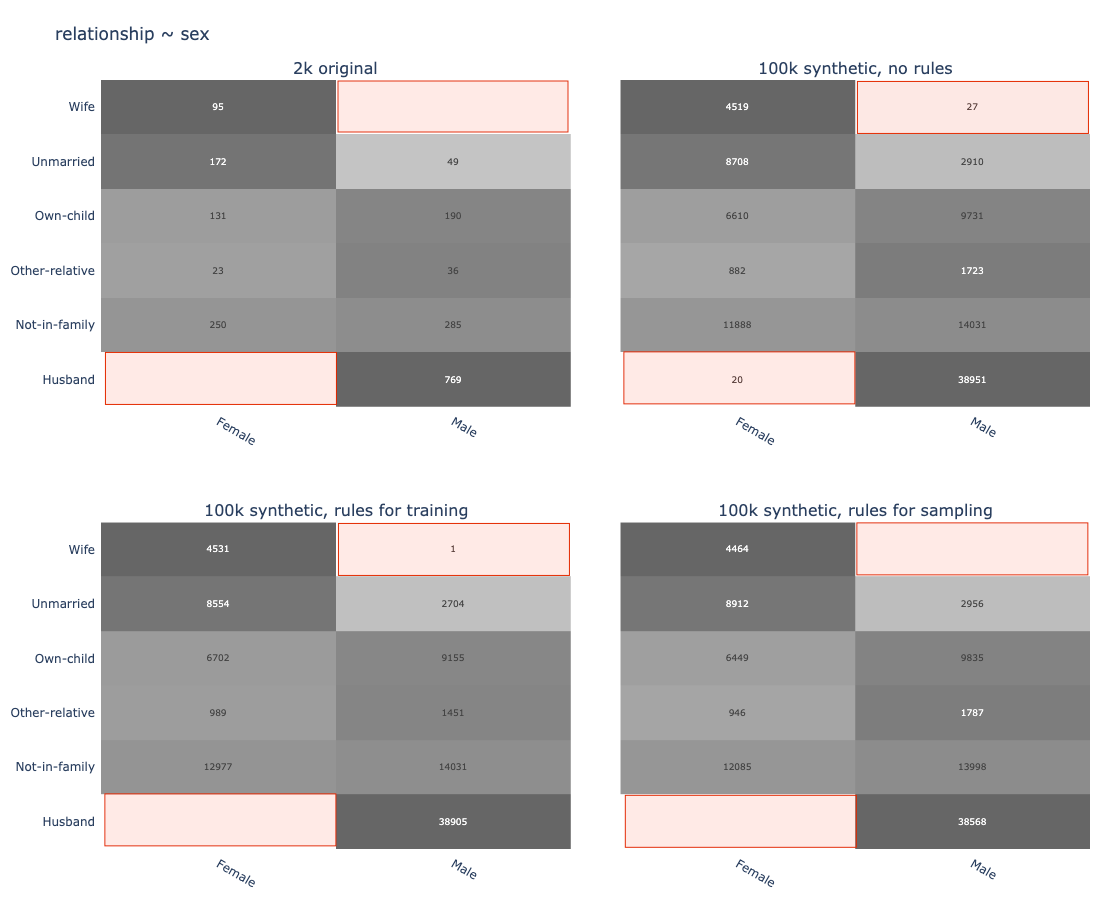}
  \caption{Rule (4) for \textit{relationship} vs. \textit{sex}}
  \label{fig:adult_gender}
\end{figure}

Figures \ref{fig:adult_age}, \ref{fig:adult_relationship}, \ref{fig:adult_education} and \ref{fig:adult_gender} display the resulting contingency tables across these experiments for each of the corresponding four rules. As one can see, the synthetic data, that doesn't explicitly know about the rules already successfully picks up and retains the signal that these combinations (highlighted in red) are rare. The average share, measured across 10 independent repeats of the experiments, of invalid records for each of the four rules is (1) 0.1\%, (2) 1.6\%, (3) 0.07\% and (4) 0.06\%. Adding the additional loss components during training, then indeed further reduces the likelihood of such invalid records to occur to (1) 0.01\%, (2) 0.8\%, (3) 0.02\% and (4) 0.003\%. Alternatively, if we enforce the rules during sampling, by setting the probabilities of invalid combinations to zero, we can effectively yield perfectly rule-adhering synthetic data. Given the much larger amount of 100k synthetic records that is being generated, these rule-adhering data sets allow any downstream consumers to infer with much greater confidence that these rules are indeed valid, when compared to an analysis that is purely informed by the original data sample of 2,000 records.

\begin{figure}[h]
  \centering
  \includegraphics[width=\linewidth]{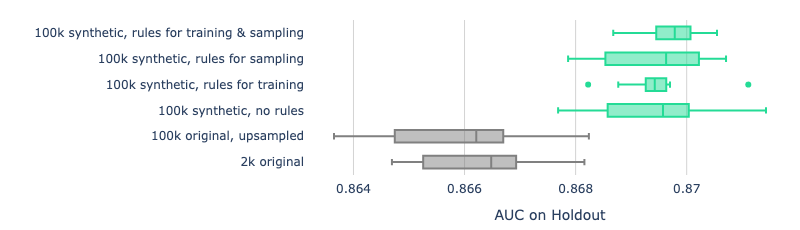}
  \caption{Model performance on the (original) holdout data}
  \label{fig:adult_aucs}
\end{figure}

Let's also assess the impact of replacing original data with synthetic data on a downstream machine learning task. For that, we pursue a train-synthetic-test-real (\textit{TSTR}) evaluation scheme \citep{esteban2017real}, and measure model performance on the remaining 46,842 holdout records, that were not part of the 2,000 sampled records, and thus also not used for the data synthesis. We then train a gradient boosting classifier\footnote{This model class was selected, as it provided the strongest out-of-the-box performance among the available sklearn models \cite{pedregosa2011scikit} for the given task.} to predict whether a person has a high \textit{income} or not. The corresponding AUCs across 10 separate runs are reported in figure~\ref{fig:adult_aucs}. As we can see, the 100,000 synthetic samples, that do not explicitly consider the rules, already yield a higher predictive performance than the original data set of 2,000 sample. Thus, for the given scenario, the generative model is a stronger learner, and can teach the downstream model by providing a large amount of diverse, yet representative synthetic samples. The incorporation of rules then does neither harm nor improve the results with respect to the learning task. We assume that this is due to the small share of invalid records in the first place, and secondly, that the predictive task at hand likely does not depend on the rule-related signals.

\section{Conclusions}

We have demonstrated the concept of rule-adhering synthetic data, and shown that it is possible to incorporate pre-existing knowledge on invalid combinations into the synthesis process. For the selected data set and the four presented rules, the rule adherence is significantly improved by adding rule-specific loss components to the training phase. Further, the rules can be perfectly satisfied by considering these during the sampling phase. The theoretically unlimited amount of rule-adhering synthetic data can then be used for any downstream data consumers, humans as well as algorithms alike, to learn the statistical information of the original data, as well as the rules by domain experts. We hope that these findings motivate further research and explorations of use cases for \textit{rule-adhering synthetic data}, as it shows promise to serve as a \textit{lingua franca} of learning.

\bibliographystyle{ACM-Reference-Format}
\bibliography{article}


\begin{thebibliography}{9}


\ifx \showCODEN    \undefined \def \showCODEN     #1{\unskip}     \fi
\ifx \showDOI      \undefined \def \showDOI       #1{#1}\fi
\ifx \showISBNx    \undefined \def \showISBNx     #1{\unskip}     \fi
\ifx \showISBNxiii \undefined \def \showISBNxiii  #1{\unskip}     \fi
\ifx \showISSN     \undefined \def \showISSN      #1{\unskip}     \fi
\ifx \showLCCN     \undefined \def \showLCCN      #1{\unskip}     \fi
\ifx \shownote     \undefined \def \shownote      #1{#1}          \fi
\ifx \showarticletitle \undefined \def \showarticletitle #1{#1}   \fi
\ifx \showURL      \undefined \def \showURL       {\relax}        \fi
\providecommand\bibfield[2]{#2}
\providecommand\bibinfo[2]{#2}
\providecommand\natexlab[1]{#1}
\providecommand\showeprint[2][]{arXiv:#2}

\bibitem[Brown et~al\mbox{.}(2020)]%
        {brown2020language}
\bibfield{author}{\bibinfo{person}{Tom Brown}, \bibinfo{person}{Benjamin Mann},
  \bibinfo{person}{Nick Ryder}, \bibinfo{person}{Melanie Subbiah},
  \bibinfo{person}{Jared~D Kaplan}, \bibinfo{person}{Prafulla Dhariwal},
  \bibinfo{person}{Arvind Neelakantan}, \bibinfo{person}{Pranav Shyam},
  \bibinfo{person}{Girish Sastry}, \bibinfo{person}{Amanda Askell},
  {et~al\mbox{.}}} \bibinfo{year}{2020}\natexlab{}.
\newblock \showarticletitle{Language models are few-shot learners}.
\newblock \bibinfo{journal}{\emph{Advances in neural information processing
  systems}}  \bibinfo{volume}{33} (\bibinfo{year}{2020}),
  \bibinfo{pages}{1877--1901}.
\newblock


\bibitem[Dua and Graff(2017)]%
        {uci}
\bibfield{author}{\bibinfo{person}{Dheeru Dua} {and} \bibinfo{person}{Casey
  Graff}.} \bibinfo{year}{2017}\natexlab{}.
\newblock \bibinfo{title}{{UCI} Machine Learning Repository}.
\newblock
\newblock
\urldef\tempurl%
\url{http://archive.ics.uci.edu/ml}
\showURL{%
\tempurl}


\bibitem[Esteban et~al\mbox{.}(2017)]%
        {esteban2017real}
\bibfield{author}{\bibinfo{person}{Crist{\'o}bal Esteban},
  \bibinfo{person}{Stephanie~L Hyland}, {and} \bibinfo{person}{Gunnar
  R{\"a}tsch}.} \bibinfo{year}{2017}\natexlab{}.
\newblock \showarticletitle{Real-valued (medical) time series generation with
  recurrent conditional gans}.
\newblock \bibinfo{journal}{\emph{arXiv preprint arXiv:1706.02633}}
  (\bibinfo{year}{2017}).
\newblock


\bibitem[Holtzman et~al\mbox{.}(2019)]%
        {holtzman2019curious}
\bibfield{author}{\bibinfo{person}{Ari Holtzman}, \bibinfo{person}{Jan Buys},
  \bibinfo{person}{Li Du}, \bibinfo{person}{Maxwell Forbes}, {and}
  \bibinfo{person}{Yejin Choi}.} \bibinfo{year}{2019}\natexlab{}.
\newblock \showarticletitle{The curious case of neural text degeneration}.
\newblock \bibinfo{journal}{\emph{arXiv preprint arXiv:1904.09751}}
  (\bibinfo{year}{2019}).
\newblock


\bibitem[Karras et~al\mbox{.}(2017)]%
        {karras2017progressive}
\bibfield{author}{\bibinfo{person}{Tero Karras}, \bibinfo{person}{Timo Aila},
  \bibinfo{person}{Samuli Laine}, {and} \bibinfo{person}{Jaakko Lehtinen}.}
  \bibinfo{year}{2017}\natexlab{}.
\newblock \showarticletitle{Progressive growing of gans for improved quality,
  stability, and variation}.
\newblock \bibinfo{journal}{\emph{arXiv preprint arXiv:1710.10196}}
  (\bibinfo{year}{2017}).
\newblock


\bibitem[Pedregosa et~al\mbox{.}(2011)]%
        {pedregosa2011scikit}
\bibfield{author}{\bibinfo{person}{Fabian Pedregosa}, \bibinfo{person}{Ga{\"e}l
  Varoquaux}, \bibinfo{person}{Alexandre Gramfort}, \bibinfo{person}{Vincent
  Michel}, \bibinfo{person}{Bertrand Thirion}, \bibinfo{person}{Olivier
  Grisel}, \bibinfo{person}{Mathieu Blondel}, \bibinfo{person}{Peter
  Prettenhofer}, \bibinfo{person}{Ron Weiss}, \bibinfo{person}{Vincent
  Dubourg}, {et~al\mbox{.}}} \bibinfo{year}{2011}\natexlab{}.
\newblock \showarticletitle{Scikit-learn: Machine learning in Python}.
\newblock \bibinfo{journal}{\emph{Journal of machine learning research}}
  \bibinfo{volume}{12}, \bibinfo{number}{Oct} (\bibinfo{year}{2011}),
  \bibinfo{pages}{2825--2830}.
\newblock


\bibitem[Platzer and Reutterer(2021)]%
        {platzer2021holdout}
\bibfield{author}{\bibinfo{person}{Michael Platzer} {and}
  \bibinfo{person}{Thomas Reutterer}.} \bibinfo{year}{2021}\natexlab{}.
\newblock \showarticletitle{Holdout-based empirical assessment of mixed-type
  synthetic data}.
\newblock \bibinfo{journal}{\emph{Frontiers in big Data}}
  (\bibinfo{year}{2021}), \bibinfo{pages}{43}.
\newblock


\bibitem[Seo et~al\mbox{.}(2021)]%
        {seo2021controlling}
\bibfield{author}{\bibinfo{person}{Sungyong Seo}, \bibinfo{person}{Sercan
  Arik}, \bibinfo{person}{Jinsung Yoon}, \bibinfo{person}{Xiang Zhang},
  \bibinfo{person}{Kihyuk Sohn}, {and} \bibinfo{person}{Tomas Pfister}.}
  \bibinfo{year}{2021}\natexlab{}.
\newblock \showarticletitle{Controlling neural networks with rule
  representations}.
\newblock \bibinfo{journal}{\emph{Advances in Neural Information Processing
  Systems}}  \bibinfo{volume}{34} (\bibinfo{year}{2021}),
  \bibinfo{pages}{11196--11207}.
\newblock


\bibitem[Tiwald et~al\mbox{.}(2021)]%
        {tiwald2021representative}
\bibfield{author}{\bibinfo{person}{Paul Tiwald}, \bibinfo{person}{Alexandra
  Ebert}, {and} \bibinfo{person}{Daniel~T Soukup}.}
  \bibinfo{year}{2021}\natexlab{}.
\newblock \showarticletitle{Representative \& fair synthetic data}.
\newblock \bibinfo{journal}{\emph{arXiv preprint arXiv:2104.03007}}
  (\bibinfo{year}{2021}).
\newblock


\end{thebibliography}

\end{document}